\documentclass{article}
\usepackage{ijcai23}

\usepackage{times}
\usepackage{microtype}
\usepackage{soul}
\usepackage{url}
\usepackage[usenames,svgnames]{xcolor}
\usepackage[colorlinks,
citecolor=NavyBlue,
linkcolor=NavyBlue,
urlcolor=NavyBlue]{hyperref}
\usepackage[utf8]{inputenc}
\usepackage{setspace}
\usepackage[small]{caption}
\usepackage{graphicx}
\usepackage{amsmath}
\usepackage{amsthm}
\usepackage{booktabs}
\usepackage{algorithm}
\usepackage{algorithmic}
\usepackage{mydef}
\usepackage{booktabs}
\usepackage{tabularx}
\usepackage{framed}
\usepackage{bbding}
\usepackage{subfloat}
\usepackage{adjustbox}
\title{A Systematic Survey of Chemical Pre-trained Models}
\author{
Jun Xia{\rm\textsuperscript{1,}}\thanks{Equal contribution. $^\dagger$\,Corresponding author.}\and
Yanqiao Zhu{\rm\textsuperscript{2,$\ast$}}\and
Yuanqi Du{\rm\textsuperscript{3,$\ast$}}\And
Stan Z. Li{\rm\textsuperscript{1,$\dagger$}}\\
\affiliations
\textsuperscript{1}Westlake University\enspace
\textsuperscript{2}University of California, Los Angeles\enspace
\textsuperscript{3}Cornell University\\
\emails
\{xiajun, stan.zq.li\}@westlake.edu.cn, yzhu@cs.ucla.edu, yd392@cs.cornell.edu
}
\begin{document}
\maketitle

\begin{abstract}
Deep learning has achieved remarkable success in learning representations for molecules, which is crucial for various biochemical applications, ranging from property prediction to drug design. However, training Deep Neural Networks (DNNs) from scratch often requires abundant labeled molecules, which are expensive to acquire in the real world. To alleviate this issue, tremendous efforts have been devoted to Chemical Pre-trained Models (CPMs), where DNNs are pre-trained using large-scale unlabeled molecular databases and then fine-tuned over specific downstream tasks.
Despite the prosperity, there lacks a systematic review of this fast-growing field. In this paper, we present the first survey that summarizes the current progress of CPMs. We first highlight the limitations of training molecular representation models from scratch to motivate CPM studies. Next, we systematically review recent advances on this topic from several key perspectives, including molecular descriptors, encoder architectures, pre-training strategies, and applications. We also highlight the challenges and promising avenues for future research, providing a useful resource for both machine learning and scientific communities.
\end{abstract}
\section{Introduction}
Extracting vector representations for molecules is critical to applying machine learning methods to a broad spectrum of molecular tasks.
Initially, molecular fingerprints are developed to encode molecules into binary vectors with rule-based algorithms~\cite{consonni2009molecular}.
Subsequently, various Deep Neural Networks (DNNs) have been employed to encode molecules in a data-driven manner. Early attempts exploit sequence-based neural architectures (e.g., RNNs, LSTMs, and transformers) to encode molecules represented in Simplified Molecular-Input Line-Entry System (SMILES) strings~\cite{weininger1988smiles}.
Later, it is argued that molecules can be naturally represented in graph structures with atoms as nodes and bonds as edges.
This inspires a line of works to leverage such structured inductive bias for better molecular representations~\cite{kearnes2016molecular}. The key advancements underneath these approaches are Graph Neural Networks (GNNs), which consider graph structures and attributive features simultaneously by recursively aggregating node features from neighborhoods~\cite{kipf2017semi-supervised}. 
Recently, another line of development of GNNs for molecular representations models 3D geometric symmetries of molecular conformations, considering the molecules are in a constant motion in 3D space by nature~\cite{schutt2017schnet}.
\begin{figure}[t]
    \begin{center}
    \includegraphics[width=1.0\linewidth]{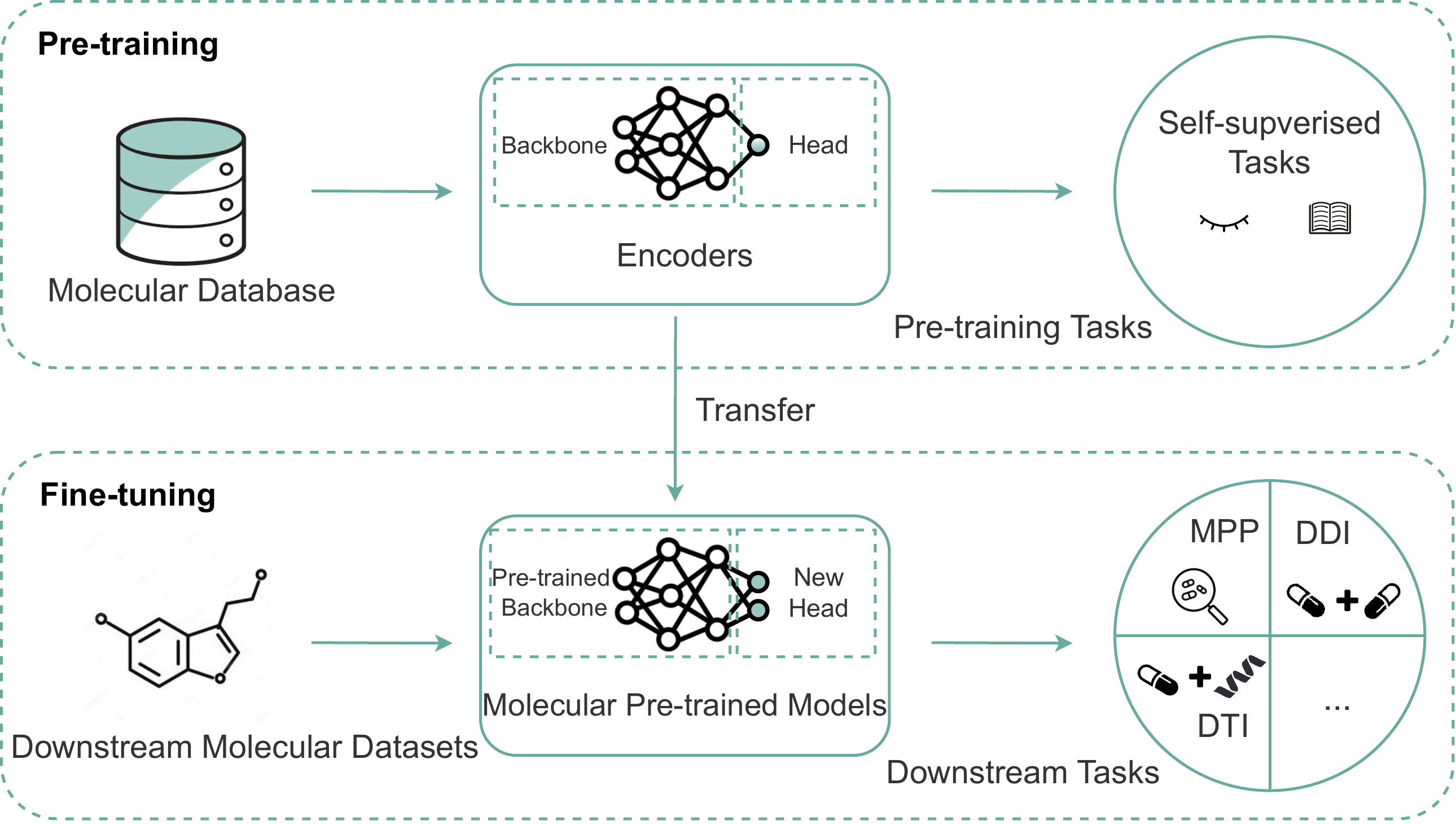}
    \end{center}
    \caption{A typical learning pipeline for Chemical Pre-trained Models (CPMs).
    MPP: Molecular Property Prediction; DDI: Drug-Drug Interactions; DTI: Drug-Target Interactions.}
   \label{fig_01}
\end{figure}
However, the majority of the above works learn molecular representations under supervised settings, which limits their widespread application in practice for the following reasons. (1) \emph{Scarcity of labeled data:} task-specific labels of molecules can be extremely scarce because molecular data labeling often requires expensive wet-lab experiments; (2) \emph{Poor out-of-distribution generalization:} learning molecules with different sizes or functional groups requires out-of-distribution generalization in many real-world cases. For example, suppose one wishes to predict the properties of a newly synthesized molecule that differs from all the previous molecules in the training set. However, models trained from scratch cannot extrapolate to out-of-distribution molecules well~\cite{Hu*2020Strategies}.
\tikzstyle{leaf}=[draw=hiddendraw,
    rounded corners,minimum height=1em,
    fill=hidden-orange!40,text opacity=1, align=center,
    fill opacity=.5,  text=black,align=left,font=\scriptsize,
    inner xsep=3pt,
    inner ysep=1pt,
    ]
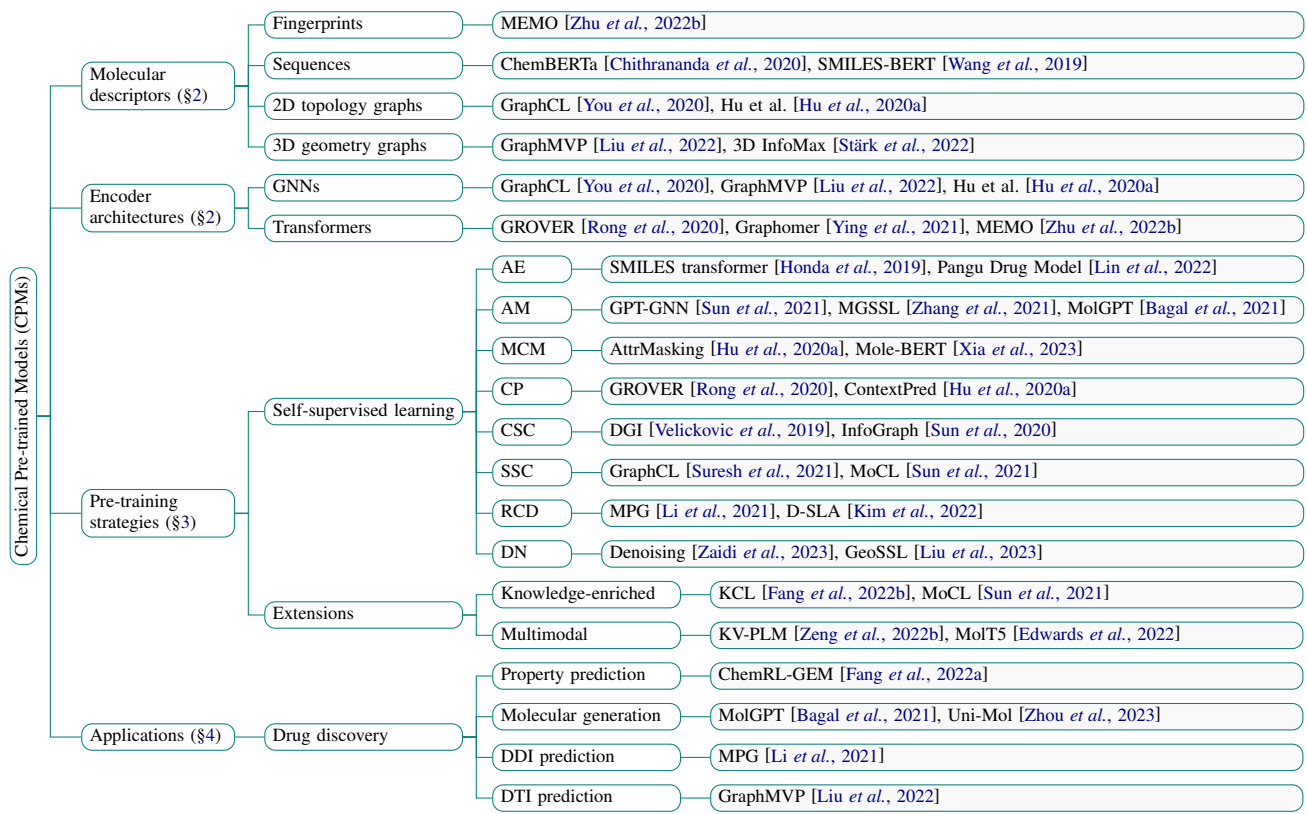
\begin{figure*}[ht]
\centering
\begin{forest}
  for tree={
  forked edges,
  grow=east,
  reversed=true,
  anchor=base west,
  parent anchor=east,
  child anchor=west,
  base=middle,
  font=\scriptsize,
  rectangle,
  draw=hiddendraw,
  rounded corners,align=left,
  minimum width=2em,
    s sep=5pt,
    inner xsep=3pt,
    inner ysep=1pt,
  },
  where level=1{text width=4.5em}{},
  where level=2{text width=6em,font=\scriptsize}{},
  where level=3{font=\scriptsize}{},
  where level=4{font=\scriptsize}{},
  where level=5{font=\scriptsize}{},
  [Chemical Pre-trained Models (CPMs),rotate=90,anchor=north,edge=hiddendraw
    [Molecular \\descriptors (\S\ref{rep}),edge=hiddendraw,text width=5.18em
        [Fingerprints, text width=6.88em, edge=hiddendraw
            [MEMO~\cite{zhu2022featurizations},leaf,text width=30.08em, edge=hiddendraw]
        ]
        [Sequences, text width=6.88em, edge=hiddendraw
            [ChemBERTa~\cite{chithrananda2020chemberta}{,} SMILES-BERT~\cite{wang2019smiles},leaf,text width=30.08em, edge=hiddendraw]
        ]
        [2D topology graphs, text width=6.88em, edge=hiddendraw
            [GraphCL~\cite{You2020GraphCL}{,} Hu et al.~\cite{Hu*2020Strategies},leaf,text width=30.08em, edge=hiddendraw]
        ]
        [3D geometry graphs, text width=6.88em, edge=hiddendraw
            [GraphMVP~\cite{liu2022pretraining}{,} 3D InfoMax~\cite{stark20213d},leaf,text width=30.08em, edge=hiddendraw]
        ]
    ]
    [Encoder\\architectures (\S\ref{rep}),edge=hiddendraw,text width=5.18em
     [GNNs, text width=6.88em, edge=hiddendraw
        [
        GraphCL~\cite{You2020GraphCL}{,} GraphMVP~\cite{liu2022pretraining}{,}
        Hu et al.~\cite{Hu*2020Strategies},leaf,text width=30.08em, edge=hiddendraw
        ]
     ]
    [Transformers, text width=6.88em, edge=hiddendraw
        [GROVER~\cite{rong2020self}{,}
        Graphomer~\cite{ying2021do}{,}
        MEMO~\cite{zhu2022featurizations}
        ,leaf,text width=30.08em, edge=hiddendraw]
    ]
    ]
    [Pre-training\\ strategies (\S\ref{PS}), edge=hiddendraw,text width=5.18em
      [Self-supervised learning,text width=6.88em, edge=hiddendraw
        [AE, edge=hiddendraw, text width=2.38em
            [SMILES transformer~\cite{honda2019smiles}{,} Pangu Drug Model~\cite{lin2022pangu},leaf,text width=25.95em, edge=hiddendraw]
        ]
        [AM, edge=hiddendraw, text width=2.38em
            [GPT-GNN~\cite{sun2021mocl}{,} MGSSL~\cite{zhang2021motif}{,} MolGPT~\cite{bagal2021molgpt},leaf,text width=25.95em, edge=hiddendraw]
        ]
        [MCM, edge=hiddendraw, text width=2.38em
            [AttrMasking~\cite{Hu*2020Strategies}{,} Mole-BERT~\cite{xia2023mole-bert},leaf,text width=25.95em, edge=hiddendraw]
        ]
        [CP, edge=hiddendraw, text width=2.38em
            [GROVER~\cite{rong2020self}{,} ContextPred~\cite{Hu*2020Strategies},leaf,text width=25.95em, edge=hiddendraw]
        ]
        [CSC, edge=hiddendraw, text width=2.38em
            [DGI~\cite{velickovic2019deep}{,} InfoGraph~\cite{Sun2020InfoGraph:},leaf,text width=25.95em, edge=hiddendraw]
        ]
        [SSC, edge=hiddendraw, text width=2.38em
            [GraphCL~\cite{suresh2021adversarial}{,}
            MoCL~\cite{sun2021mocl},leaf,text width=25.95em, edge=hiddendraw]
        ]
        [RCD,edge=hiddendraw, text width=2.38em
            [MPG~\cite{li2021effective}{,} D-SLA~\cite{DBLP:journals/corr/abs-2202-02989},leaf,text width=25.95em, edge=hiddendraw]
        ]
        [DN, edge=hiddendraw, text width=2.38em
            [Denoising~\cite{zaidi2022pre}{,} GeoSSL~\cite{liu2023molecular},leaf,text width=25.95em, edge=hiddendraw]
        ]
      ]
    [Extensions,text width=6.88em, edge=hiddendraw
      [Knowledge-enriched, text width=6.5em, edge=hiddendraw
        [
        KCL~\cite{fang2021molecular}{,}
        MoCL~\cite{sun2021mocl}
        ,leaf,text width=21.8em, edge=hiddendraw]
      ]
      [Multimodal,text width=6.5em, edge=hiddendraw
        [KV-PLM~\cite{zeng2022deep}{,} MolT5~\cite{edwards2022translation},leaf,text width=21.8em, edge=hiddendraw]
      ]
    ]
    ]
   [Applications (\S\ref{app}), edge=hiddendraw,text width=5.18em
       [Drug discovery,text width=6.88em, edge=hiddendraw
            [Property prediction,text width=6.5em, edge=hiddendraw
            [
            ChemRL-GEM~\cite{Fang2021geo}
            ,leaf,text width=21.8em, edge=hiddendraw]
            ]
            [Molecular generation,text width=6.5em, edge=hiddendraw
            [
            MolGPT~\cite{bagal2021molgpt}{,} Uni-Mol~\cite{zhou2022uni},leaf,text width=21.8em, edge=hiddendraw]
            ]
          [DDI prediction,text width=6.5em, edge=hiddendraw
            [MPG~\cite{li2021effective},leaf,text width=21.8em, edge=hiddendraw]
          ]
         [DTI prediction,text width=6.5em, edge=hiddendraw
            [GraphMVP~\cite{liu2022pretraining},leaf,text width=21.8em, edge=hiddendraw]
          ]
       ]
    ]
  ]
\end{forest}
\caption{A taxonomy of Chemical Pre-trained Models (CPMs) with representative examples.}
\label{taxonomy_of_pGMs}
\end{figure*}

Pre-trained Language Models (PLMs) have been a potential solution to the above challenges in the Natural Language Processing (NLP) community~\cite{devlin2019bert}. Inspired by their success, as shown in Fig.~\ref{fig_01}, Chemical Pre-trained Models (CPMs) have been introduced to learn universal molecular representations from massive unlabeled molecules and then fine-tuned over specific downstream tasks. Initially, researchers adopt sequence-based pre-training strategies on string-based molecular data such as SMILES. A typical strategy is to pre-train the neural encoders to predict randomly masked tokens like BERT~\cite{devlin2019bert}.
This line of works include ChemBERTa~\cite{chithrananda2020chemberta}, SMILES-BERT~\cite{wang2019smiles}, Molformer~\cite{ross2022molformer}, etc.
More recently, the community explores pre-training on (both 2D and 3D) molecular graphs. 
For example, \cite{Hu*2020Strategies} propose to mask atom or edge attributes and predict the masked attributes. \cite{liu2022pretraining} pre-train GNNs via maximizing the correspondence between 2D topological and 3D geometric structures.

Although CPMs have been increasingly applied in molecular representation learning, this rapidly expanding field still lacks a systematic review. In this paper, we present the first survey for CPMs to assist audiences of diverse backgrounds in understanding, using, and developing CPMs for various practical tasks.
The contributions of this work can be summarized from the following four aspects.
\textbf{(1)} \emph{A structured taxonomy.}
    A broad overview of the field is presented with a structured taxonomy that categorizes existing works from four perspectives (Fig.~\ref{taxonomy_of_pGMs}): molecular descriptors, encoder architectures, pre-training strategies, and applications.
\textbf{(2)} \emph{Thorough review of the current progress.}
    Based on the taxonomy, the current research progress of pre-trained models for molecules is systematically delineated.
\textbf{(3)} \emph{Abundant additional resources.}
    Abundant resources including open-sourced CPMs, available datasets, and an important paper list are collected and can be found at \url{https://github.com/junxia97/awesome-pretrain-on-molecules}. These resources will be continuously updated on a regular basis.
\textbf{(4)} \emph{Discussion of future directions.}
    The limitations of existing works are discussed and several promising research directions are highlighted.

\section{Molecular Descriptors and Encoders}
\label{rep}
In order to feed molecules to DNNs, molecules have to be featurized in numerical descriptors. Various descriptors are designed to describe molecules in a concise format. In this section, we briefly review these molecular descriptors and their corresponding neural encoder architectures.

\textbf{Fingerprints (FP).}
Molecular fingerprints describe the presence or absence of particular substructures of a molecule with binary strings. For example, PubChemFP~\cite{wang2017pubchem} encodes 881 structural key types that correspond to the substructures for a fragment of compounds in the PubChem database. 

\textbf{Sequences.}
The most frequently used sequential descriptor for molecules is the Simplified Molecular-Input Line-Entry System (SMILES)~\cite{weininger1988smiles} owing to its versatility and interpretability. Each atom is represented as a respective ASCII symbol. Chemical bonds, branching, and stereochemistry are denoted by specific symbols.
Transformers~\cite{vaswani2017attention} are a powerful neural model for processing sequences and modeling the complex relationships among each token. We can split sequence-based molecular descriptors into a series of tokens denoting atoms/bonds at first and then apply transformers on top of these tokens~\cite{chithrananda2020chemberta,wang2019smiles}.

\textbf{2D graphs.}
Molecules can be represented as 2D graphs naturally, with atoms as nodes and bonds as edges. Each node and edge can also carry feature vectors denoting the atom types/chirality and bond types/direction for instance~\cite{Hu*2020Strategies}. Here, GNNs~\cite{kipf2017semi-supervised,xu2018how} can be used to learn 2D molecular graph representations. Some hybrid architectures of GNNs and transformers~\cite{rong2020self,ying2021do} can also be leveraged to capture the topological structures of molecular graphs.
\textbf{3D graphs.}
3D geometries represent the spatial arrangements of atoms of the molecule in the 3D space, where each atom is associated with its type and coordinate plus some optional geometric attributes such as velocity.
The advantage of using 3D geometry is that the conformer information 
is critical to many molecular properties, especially quantum properties. In addition, it is also possible to directly leverage stereochemistry information such as chirality given the 3D geometries.
A number of approaches~\cite{schutt2017schnet,satorras2021n,du2022se} have developed message-passing mechanisms on 3D geometries, which enable the graph representations to follow certain physical symmetries, such as equivariance to translations and rotations.

\begin{figure*}[ht]
    \centering
    \subfloat[Autoencoding]
  {
  \label{fig1-a}
    \includegraphics[scale=0.54]{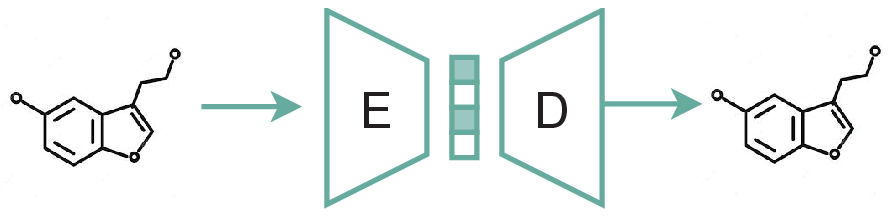}
  }\qquad
    \subfloat[Autoregressive Modeling]
  {
    \label{fig1-b}
    \includegraphics[scale=0.54]{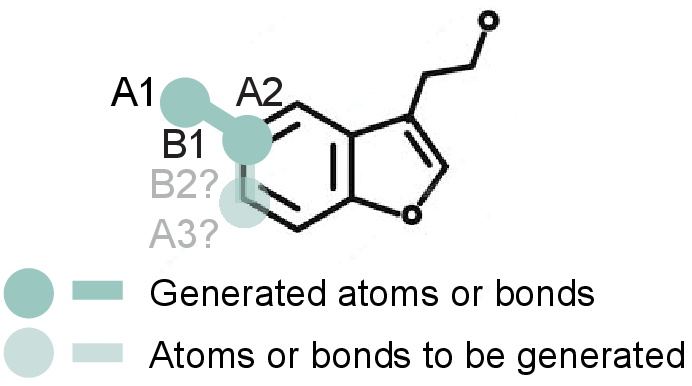}
  }\qquad
    \subfloat[Masked Components Modeling]
  {
    \label{fig1-c}
    \includegraphics[scale=0.54]{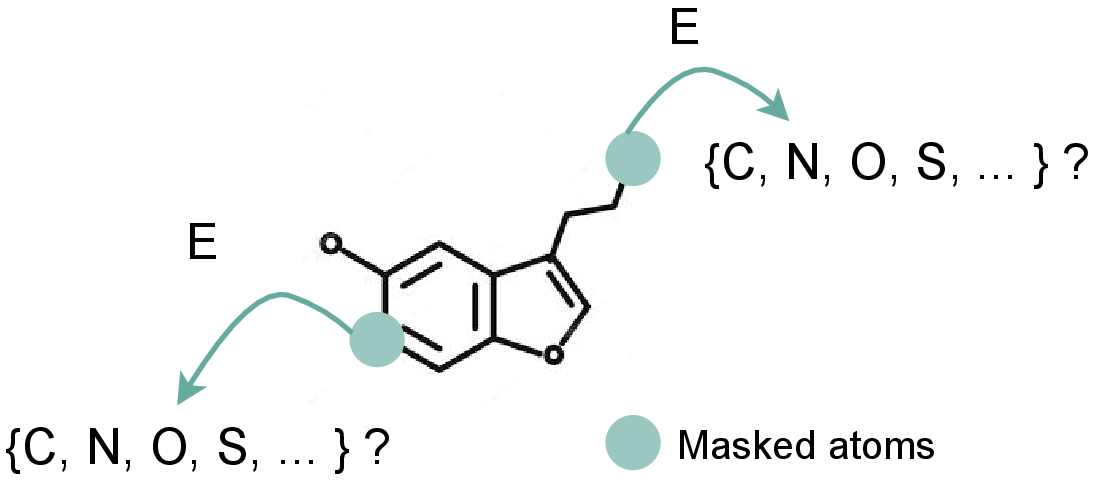}
  }\\
    \subfloat[Context Prediction]
  {
    \label{fig1-d}
    \includegraphics[scale=0.54]{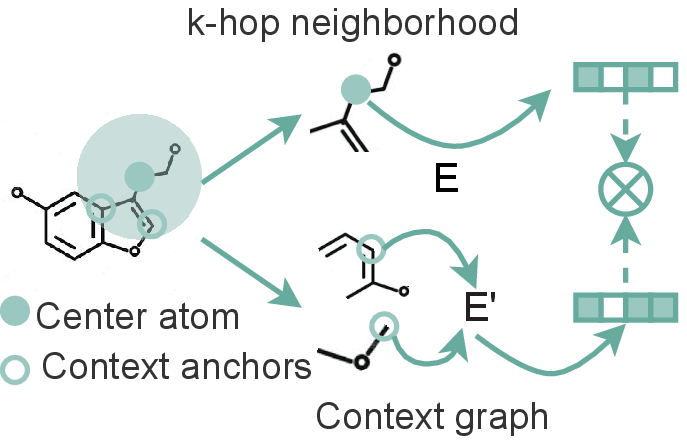}
  }
    \subfloat[Contrastive Learning]
  {
    \label{fig1-e}
    \includegraphics[scale=0.54]{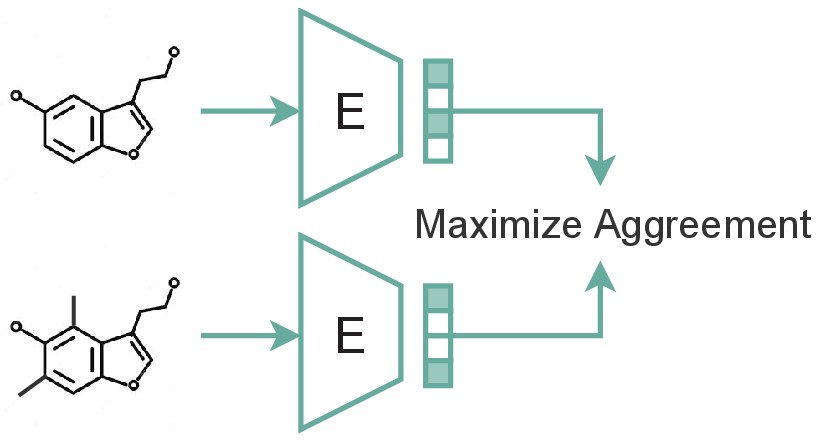}
  }
    \subfloat[Replaced Component Detection]
  {
    \label{fig1-f}
    \includegraphics[scale=0.54]{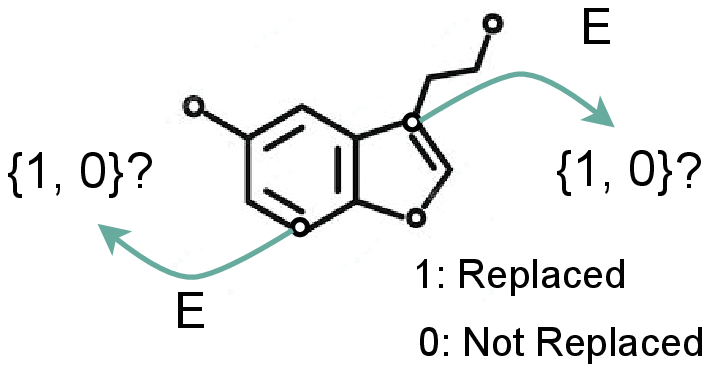}
  }
    \subfloat[Denoising]
  {
    \label{fig1-g}
    \includegraphics[scale=0.54]{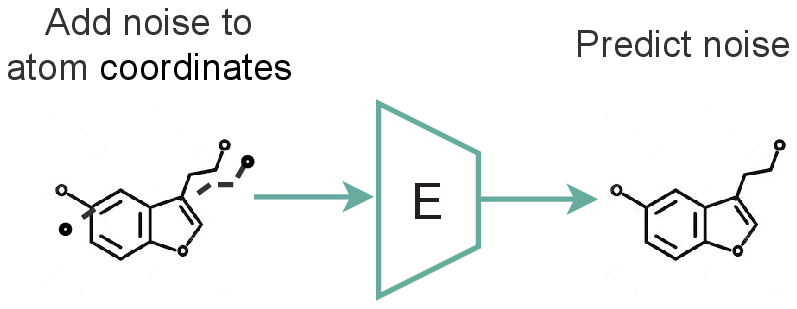}
  }
    \caption{Semantic diagrams of seven unsupervised pre-training strategies. E: encoder; D: decoder.}
    \label{fig_strategy}
\end{figure*}
\section{Pre-training Strategies}
\label{PS}
In this section, we elaborate on several representative self-supervised pre-training strategies of CPMs.
\subsection{AutoEncoding (AE)}
Reconstructing molecules with autoencoders (Fig.~\ref{fig1-a}) serves as a natural self-supervised target for learning expressive molecular representations. The prediction in molecule reconstructions is (partial) structures of the given molecules such as the attributes of a subset of atoms or chemical bonds.
A typical example is SMILES transformer~\cite{honda2019smiles}, which leverages a transformer-based encoder-decoder network and learns the representations by reconstructing molecules represented by SMILES strings. 
More recently, unlike conventional autoencoders with the same types for the input and output data, \cite{lin2022pangu} pre-train a graph-to-sequence asymmetric conditional variational autoencoder to learn molecular representations.
Although autoencoders can learn meaningful representations for molecules, they focus on single molecules and fail to capture inter-molecule relationships, which limits their performance in some downstream tasks~\cite{li2021effective}.

\subsection{Autoregressive Modeling (AM)}
\label{GAM}
Autoregressive Modeling (AM) factorizes the molecular input as a sequence of sub-components and then it predicts the sub-components one by one, conditioned on previous sub-components in the sequence. Following the idea of GPT~\cite{brown2020language} in NLP, MolGPT~\cite{bagal2021molgpt} pre-trains a transformer network to predict the next token in the SMILES strings in such an autoregressive manner.
For molecular graphs, GPT-GNN~\cite{hu2020gpt} reconstructs the molecular graph in a sequence of steps (Fig.~\ref{fig1-b}), in contrast to graph autoencoders that reconstruct the whole graph at once.
In particular, given a graph with its nodes and edges randomly masked, GPT-GNN generates one masked node and its edges at a time and maximizes the likelihood of the node and edges generated in each iteration. Then, it iteratively generates nodes and edges until all masked nodes are generated. Analogously, MGSSL~\cite{zhang2021motif} generates molecular graph motifs instead of individual atoms or bonds autoregressively. Formally, such autoregressive modeling objectives can be written as
\begin{equation}
    \mathcal{L}_{\text{AM}}=-\mathbb{E}_{\mathcal{M}\in \mathcal{D}}\sum_{i=1}^{|\mathcal{C}|} \log  p(\mathcal{C}_{i} \mid \mathcal{C}_{<i}),
\end{equation}
where $\mathcal{C}_{i},\mathcal{C}_{<i}$ are the attributes of $i$-th component and the ones generated before the index $i$ in the molecule $\mathcal{M}$, respectively. Compared with other strategies, AM allows CPMs to perform better at generating molecules its training procedure resembles that of molecule generation~\cite{bagal2021molgpt}. However, AM is more computationally expensive and requires a preset ordering of atoms or bonds beforehand, which may be inappropriate for molecules because the atoms or bonds do not present inherent orders.
\subsection{Masked Component Modeling (MCM)}
\label{MCM}
In language domains, Masked Language Modeling (MLM) has emerged as a dominant pre-training objective. Specifically, MLM randomly masks out tokens from the input sentences, where the model can be trained to predict those masked tokens using the remaining tokens~\cite{devlin2019bert}.
Masked Component Modeling (MCM, Fig.~\ref{fig1-c}) generalizes the idea of MLM for molecules.
Specifically, MCM masks out some components (e.g., atoms, bonds, and fragments) of the molecules and then trains the model to predict them given the remaining components. Generally, its objective can be formulated as
\begin{equation}
    \mathcal{L}_{\text{MCM}}=-\mathbb{E}_{\mathcal{M}\in \mathcal{D}}\sum_{\widetilde{\mathcal{M}} \in m(\mathcal{M})} \log p(\widetilde{\mathcal{M}} \mid \mathcal{M}\backslash m(\mathcal{M})),
\end{equation}
where $m(\mathcal{M})$ denotes the masked components from the molecule $\mathcal{M}$ and $\mathcal{M}\backslash m(\mathcal{M})$ are the remaining components. For sequence-based pre-training, ChemBERTa~\cite{chithrananda2020chemberta}, SMILES-BERT~\cite{wang2019smiles}, and Molformer~\cite{ross2022molformer} mask random characters in the SMILES strings and then recover them based on the output of the transformer of the corrupted SMILES strings.
For molecular graph pre-training, \cite{Hu*2020Strategies} propose to randomly mask input atom/chemical bond attributes and pre-train the GNNs to predict them.
Similarly, GROVER~\cite{rong2020self} attempts to predict the masked subgraphs to capture the contextual information in the molecular graphs. Recently, Mole-BERT~\cite{xia2023mole-bert} argues that masking the atom types could be problematic due to the extremely small and unbalanced atom set in nature. To mitigate this issue, they design a context-aware tokenizer to encode atoms as chemically meaningful discrete values for masking.

MCM is especially beneficial for richly-annotated molecules. For example, masking atom attributes enables GNNs to learn simple chemistry rules such as valency, as well as potentially other complex chemistry descriptors such as the electronic or steric effects of functional groups. Additionally, compared with the aforementioned AM strategies, MCM predicts the masked components based on their surrounding environments as opposed to AM which merely relies on preceding components in a predefined sequence. Hence, MCM can capture more complete chemical semantics. However, as MCM often masks a fixed portion of each molecule during pre-training following BERT~\cite{devlin2019bert}, it cannot train on all the components in each molecule, which results in less efficient sample utilization.

\subsection{Context Prediction (CP)}
Context Prediction (CP, Fig.~\ref{fig1-d}) aims to capture the semantics of molecules/atoms in an explicit, context-aware manner. Generally, CP can be formulated as
\begin{equation}
    \mathcal{L}_{\text{CP}}=-\mathbb{E}_{\mathcal{M}\in \mathcal{D}}\log p\left(t \mid \mathcal{M}_1, \mathcal{M}_2\right),
\end{equation}
where $t=1$ if neighborhood components $\mathcal{M}_1$ and surrounding contexts $\mathcal{M}_2$ share the same center atom and otherwise $t=0$. For example, \cite{Hu*2020Strategies} use a binary classification of whether the subgraphs in molecules and surrounding context structures belong to the same node. 
While simple and effective, CP requires an auxiliary neural model to encode the context into a fixed vector, adding extra computational overhead for large-scale pre-training.

\subsection{Contrastive Learning (CL)}
Contrastive Learning (CL, Fig.~\ref{fig1-e}) pre-trains the model by maximizing the agreement between a pair of similar inputs, such as two different augmentations or descriptors of the same molecule. According to the contrastive granularity (e.g., molecule- or substructure-level), we introduce two categories of CL in CPMs: Cross-Scale Contrast (CSC) and Same-Scale Contrast (SSC).

\paragraph{Cross-Scale Contrast (CSC).}
Deep InfoMax is a representative CSC model that is originally proposed for learning image representations by contrasting a pair of an image and its local regions against other negative pairs~\cite{hjelm2018learning}. For molecular graphs, InfoGraph~\cite{Sun2020InfoGraph:} follows this idea by contrasting molecule- and substructure-level representations, which can be formally described as
\begin{equation}
    \mathcal{L}_{\text{CSC}}=-\mathbb{E}_{\mathcal{M}\in \mathcal{D}}\left[\log s(\mathcal{M}, \mathcal{C})-\log \sum_{\mathcal{C}^{-} \in \mathcal{N}} s(\mathcal{M},\mathcal{C}^{-})\right],
\end{equation}
where $\mathcal{N}$ is a set of negative samples, $\mathcal{C}$ is a substructure of $\mathcal{M}$, $\mathcal{C}^{-}$ is a substructure of the other molecule, and $s(\cdot,\cdot)$ denotes a similarity metric.
Follow-up work MVGRL~\cite{hassani2020contrastive} performs node diffusion to generate an augmented molecular graph and then maximizes the similarity between original and augmented views by contrasting atom representations of one view with molecular representations of the other view and vice versa.
\paragraph{Same-Scale Contrast (SSC).}
Same-Scale Contrast (SSC) performs contrastive learning on individual molecules by pushing the augmented molecule close to the anchor molecule (positive pairs) and away from other molecules (negative pairs). For example, GraphCL~\cite{You2020GraphCL} and its variants~\cite{you2021graph,sun2021mocl,suresh2021adversarial,xu2021infogcl,fang2021molecular,wang2021molecular,10.1145/3485447.3512156,wang2022improving} propose various augmentation strategies for molecule-level pre-training represented by graphs. 
Additionally, some recent works maximize the agreement between various descriptors of identical molecules and repel the different ones. For example, SMICLR~\cite{pinheiro2022smiclr} jointly leverages a graph encoder and a SMILES string encoder to perform SSC; MM-Deacon~\cite{guo2022multilingual} utilizes two separate transformers to encode the SMILES and the International Union of Pure and Applied Chemistry (IUPAC) of molecules, after which a contrastive objective is used to promote similarity of SMILES and IUPAC representations from the same molecule; 3DInfoMax~\cite{stark20213d} proposes to maximize the agreements between the learned 3D geometry and 2D graph representations; GeomGCL~\cite{li2021geomgcl} adopts a dual-view Geometric Message Passing Neural Network (GeomMPNN) to encode both 2D and 3D graphs of a molecule and design a geometric contrastive objective. The general formulation of the SSC pre-training objective is
\begin{equation}
    \mathcal{L}_{\text{SSC}}=-\mathbb{E}_{\mathcal{M}\in \mathcal{D}}\left[\log s(\mathcal{M}, \mathcal{M}^{\prime})-\log \sum_{\mathcal{M}^{-} \in \mathcal{N}} s(\mathcal{M},\mathcal{M}^{-})\right],
\end{equation}
where $\mathcal{M}^{\prime}$ is the augmented versions or other descriptors of the molecule $\mathcal{M}$, and $\mathcal{N}$ is the set of negative samples.

Although CL has achieved promising results, several critical issues impede its broader applications. Firstly, it is difficult to preserve semantics during molecular augmentations. Existing solutions pick augmentations with manual trial-and-errors~\cite{You2020GraphCL}, cumbersome optimization~\cite{you2021graph}, or through the guidance of expensive domain knowledge~\cite{sun2021mocl}, but an efficient and principled way to design the chemically appropriate augmentation for molecular pre-training is still lacking. Also, the assumption behind CL that pulls similar representations closer may not always hold true for molecular representation learning. For example, in the case of molecular activity cliffs~\cite{stumpfe2019evolving}, similar molecules hold completely different properties. Therefore, it remains unsolved which pre-training strategies can better capture discrepancies between molecules. Additionally, the CL objective in most CPMs randomly chooses all other molecules in one batch as the negative samples regardless of their true semantics, which will undesirably repel the molecules of similar properties and undermine the performance due to the false negatives~\cite{xia2022progcl}.

\subsection{Replaced Components Detection (RCD)}
Replaced Components Detection (RCD, Fig.~\ref{fig1-f}) proposes to recognize randomly replaced components of the input molecules. For example, MPG~\cite{li2021effective} splits each molecule into two parts, shuffles their structure by combining parts from two molecules, and trains the encoder to detect whether the combined parts belong to the same molecule.
This objective can be written as
\begin{equation}
   \mathcal{L}_{\text{RCD}}=-\mathbb{E}_{\mathcal{M}\in \mathcal{D}}\left[\log p(t \mid \mathcal{M}_1, \mathcal{M}_2)\right],
\end{equation}
where $t=1$ if the two parts $\mathcal{M}_1$ and $\mathcal{M}_2$ are from the same molecule $\mathcal{M}$ and $t = 0$ otherwise. 
While RCD can uncover intrinsic patterns in molecular structures, the encoders are pre-trained to always produce the same ``non-replacement'' label for all natural molecules and ``replacement'' label for randomly combined molecules. However, in downstream tasks, the input molecules are all natural ones, causing the molecular representations produced by RCD to be less distinguishable.


\subsection{DeNoising (DN)}
Inspired by the success of denoising diffusion probabilistic models~\cite{ho2020denoising}, DeNoising (DN, Fig.~\ref{fig1-g}) has been recently adopted as a pre-training strategy for learning molecular representations as well. A recent work~\cite{zaidi2022pre} adds noise to atomic coordinates of 3D molecular geometry and pre-trains the encoders to predict the noise. They demonstrate that such a denoising objective approximates learning a molecular force field. A concurrent work, Uni-Mol~\cite{zhou2022uni} adds noise to atomic coordinates motivated by the fact that masked atom types can be easily inferred given 3D atomic positions. More recently, GeoSSL~\cite{liu2023molecular} proposes a distance denoising pre-training method to model the dynamic nature of 3D molecules. Generally, the pre-training objective of denoising can be formulated as
\begin{equation}
   \mathcal{L}_{\text{DN}}=\mathbb{E}_{\mathcal{M}\in \mathcal{D}}\|\epsilon - f_\theta(\widetilde{\mathcal{M}})\|^2,
\end{equation}
where $\epsilon$ denotes the added noise, $\widetilde{\mathcal{M}}$ denotes the input molecule $\mathcal{M}$ with noise added, and $f_\theta(\cdot)$ denotes the encoders that predict the noise. 
\subsection{Extensions}
\paragraph{Knowledge-enriched pre-training.}
CPMs usually learn general molecular representations from a large molecular database. However, they often lack domain-specific knowledge. To improve their performance, several recent works try to inject external knowledge into CPMs. For example, GraphCL~\cite{You2020GraphCL} first points out that bond perturbations (adding or dropping the bonds as data augmentations) are conceptually incompatible with domain knowledge and empirically not helpful for contrastive pre-training on chemical compounds. Therefore, they avoid adopting bond perturbations for molecular graph augmentation. More explicitly, MoCL~\cite{sun2021mocl} proposes a domain knowledge-based molecular augmentation operator called substructure substitution, in which a valid substructure of a molecule is replaced by a bioisostere which produces a new molecule with similar physical or chemical properties as the original one. More recently, KCL~\cite{fang2021molecular} constructs a chemical element Knowledge Graph (KG) to summarize microscopic associations between chemical elements and presents a novel Knowledge-enhanced Contrastive Learning (KCL) framework for molecular representation learning. Additionally, MGSSL~\cite{zhang2021motif} first leverages existing algorithms~\cite{degen2008art} to extract semantically meaningful motifs and then pre-trains neural encoders to predict the motifs in an autoregressive manner. ChemRL-GEM~\cite{Fang2021geo} proposes to utilize molecular geometry information to enhance molecular graph pre-training. It designs a geometry-based GNN architecture as well as several geometry-level self-supervised learning strategies (the bond lengths prediction, the bond angles prediction, and the atomic distance matrices prediction) to capture the molecular geometry knowledge during pre-training. Although knowledge-enriched pre-training helps CPMs capture chemical domain knowledge, it requires expensive prior knowledge as guidance, which poses a hurdle to broader applications when the prior is incomplete, incorrect, or expensive to obtain.
\begin{table*}[t]
\caption{A summary of representative Chemical Pre-trained Models (CPMs) in literature.}
\label{Table_pGMs}
\setlength{\tabcolsep}{2.8pt}
\centering
\begin{adjustbox}{max width=\linewidth}
\begin{tabular}{lllllllc}
\toprule 
& \textbf{Model} & \textbf{Input} & \textbf{Backbone architecture} &\textbf{Pre-training task} & \textbf{Pre-training database}   & \textbf{\#Params.} & \textbf{Link}  \\
\midrule
\multirow{4}{*}{\rotatebox{90}{Sequence}}  
 &SMILES Transformer~\cite{honda2019smiles}  & SMILES &Transformer & AE & ChEMBL (861K)~\cite{gaulton2012chembl}& --- &\href{https://github.com/DSPsleeporg/smiles-transformer}{Link}\\
 &ChemBERTa~\cite{chithrananda2020chemberta}  & SMILES/SELFIES~\cite{krenn2020self} &Transformer & MCM & PubChem (77M)~\cite{wang2017pubchem} & ---&\href{https://github.com/seyonechithrananda/bert-loves-chemistry}{Link}\\
&SMILES-BERT~\cite{wang2019smiles}  & SMILES &Transformer &MCM &ZINC15 ($\sim$18.6M)~\cite{zinc15} &--- &\href{https://github.com/uta-smile/SMILES-BERT}{Link}\\
&Molformer~\cite{ross2022molformer}  & SMILES &Transformer & MCM &ZINC15 (1B) + PubChem (111M) & ---& ---\\
\midrule
\multirow{18}{*}{\rotatebox{90}{Graph / Geometry}}  
&Hu et al.~\cite{Hu*2020Strategies}  &Graph&5-layer GIN  & CP + MCM & ZINC15 (2M) + ChEMBL (456K) &$\sim$2M &\href{https://github.com/snap-stanford/pretrain-gnns}{Link}\\
&GraphCL~\cite{You2020GraphCL}  &Graph &5-layer GIN &SSC  & ZINC15 (2M) & $\sim$2M&\href{https://github.com/Shen-Lab/GraphCL}{Link} \\
&JOAO~\cite{you2021graph} &Graph  &5-layer GIN  &SSC  & ZINC15 (2M)& $\sim$2M&\href{https://github.com/Shen-Lab/GraphCL_Automated}{Link}\\
&AD-GCL~\cite{suresh2021adversarial} &Graph  & 5-layer GIN &SSC  & ZINC15 (2M) & $\sim$2M &\href{https://github.com/susheels/adgcl}{Link}\\
&GraphLoG~\cite{xu2021self} &Graph  & 5-layer GIN &SSC  & ZINC15 (2M) &$\sim$2M&\href{https://github.com/DeepGraphLearning/GraphLoG}{Link} \\
&MGSSL~\cite{zhang2021motif}&Graph &5-layer GIN  &MCM + AM  & ZINC15 (250K) &$\sim$2M&\href{https://github.com/zaixizhang/MGSSL}{Link} \\
&MPG~\cite{li2021effective} &Graph  &MolGNet~\cite{li2021effective} & RCD + MCM & ZINC + ChEMBL (11M)  & 53M&\href{https://github.com/pyli0628/MPG.git}{Link} \\
&LP-Info~\cite{you2022bringing} &Graph  &5-layer GIN &SSC  & ZINC15 (2M) & $\sim$2M&\href{https://github.com/Shen-Lab/GraphCL_Automated}{Link} \\
&SimGRACE~\cite{10.1145/3485447.3512156} &Graph &5-layer GIN&SSC  & ZINC15 (2M) & $\sim$2M&\href{https://github.com/junxia97/SimGRACE}{Link}\\
&GraphMAE~\cite{hou2022graphmae} &Graph &5-layer GIN&AE  & ZINC15 (2M) & $\sim$2M&\href{https://github.com/THUDM/GraphMAE}{Link}\\
&MGMAE~\cite{DBLP:conf/cikm/FengWLDWX22} &Graph &5-layer GIN&AE  & ZINC15 (2M) + ChEMBL (456K) & $\sim$2M&---\\
&GROVER~\cite{rong2020self} &Graph & GTransformer~\cite{rong2020self}& CP + MCM  & ZINC + ChEMBL (10M)  & 48M$\sim$100M&\href{https://github.com/tencent-ailab/grover}{Link}  \\
&MolCLR~\cite{Wang2021MolCLRMC} &Graph & GCN + GIN & SSC & PubChem (10M) &---&\href{https://github.com/yuyangw/MolCLR}{Link}\\
&Graphomer~\cite{ying2021do}&Graph &Graphomer~\cite{ying2021do} &Supervised & PCQM4M-LSC ($\sim$3.8M)~\cite{DBLP:conf/nips/HuFRNDL21}& --- & \href{https://github.com/microsoft/Graphormer}{Link}\\
&3D-EMGP~\cite{jiao2023energy} & Geometry & E(3)-equivariant GNNs & DN & GEOM (100K)~\cite{DBLP:journals/corr/abs-2006-05531}  & --- &\href{https://github.com/jiaor17/3D-EMGP}{Link}\\
&Mole-BERT~\cite{xia2023mole-bert} &Graph &5-layer GIN&MCM + SSC  & ZINC15 (2M) & $\sim$2M&\href{https://github.com/junxia97/Mole-BERT}{Link}\\
&Denoising~\cite{zaidi2022pre} & Geometry & GNS~\cite{sanchez2020learning} & DN & PCQM4Mv2 ($\sim$3.4M)  & --- &\href{https://github.com/shehzaidi/pre-training-via-denoising}{Link}\\
&GeoSSL~\cite{liu2023molecular} & Geometry &PaiNN~\cite{DBLP:conf/icml/SchuttUG21} & DN & Molecule3D~\cite{DBLP:journals/corr/abs-2110-01717} ($\sim$1M) & --- & \href{https://github.com/chao1224/GeoSSL}{Link} \\
\midrule
\multirow{14}{*}{\rotatebox{90}{Multimodal / External knowledge}} 
&DMP~\cite{Zhu2021DualviewMP} &Graph + SMILES & DeeperGCN + Transformer & MCM + SSC & PubChem (110M) &104.1M&\href{https://github.com/dual-view-molecule-pretraining/dmp}{Link}\\
&GraphMVP~\cite{liu2022pretraining} & Graph + Geometry &5-layer GIN + SchNet~\cite{schutt2017schnet} & SSC + AE & GEOM (50K)  & $\sim$2M & \href{https://github.com/chao1224/GraphMVP}{Link} \\
&3D Infomax~\cite{stark20213d} & Graph + Geometry &  PNA~\cite{corso2020principal}& SSC & QM9 (50K) + GEOM (140K) + QMugs (620K)  & ---&\href{https://github.com/HannesStark/3DInfomax}{Link}\\
&KCL~\cite{fang2021molecular} & Graph + Knowledge Graph  & GCN + KMPNN~\cite{fang2021molecular} &SSC  & ZINC15 (250K) & \textless 1M&\href{https://github.com/ZJU-Fangyin/KCL}{Link}\\
&KV-PLM~\cite{zeng2022deep} & SMILES + Text &Transformer &MLM + MCM &PubChem (150M)& $\sim$110M&\href{https://github.com/thunlp/KV-PLM}{Link}\\
&MEMO~\cite{zhu2022featurizations} & SMILES + FP + Graph + Geometry & Transformer + GIN + SchNet & SSC &GEOM (50K)&---&---\\
&MolT5~\cite{edwards2022translation} & SMILES + Text &Transformer &Replace Corrupted Spans & ZINC-15 (100M) & 60M / 770M &\href{https://github.com/blender-nlp/MolT5}{Link}\\
&MICER~\cite{yi2022micer} & SMILES + Image &CNNs + LSTM &AE &ZINC20& --- & \href{https://github.com/Jiacai-Yi/MICER}{Link}\\
&MM-Deacon~\cite{guo2022multilingual} & SMILES + IUPAC &Transformer &SSC &PubChem& 10M & ---\\
&PanGu Drug Model~\cite{lin2022pangu} & Graph + SELFIES~\cite{krenn2020self} &Transformer & AE &ZINC20 + DrugSpaceX + UniChem ($\sim$1.7B)  & $\sim$104M  & \href{http://pangu-drug.com/}{Link}\\
&KPGT~\cite{DBLP:conf/kdd/LiZZ22} & SMILES + FP & LiGhT~\cite{DBLP:conf/kdd/LiZZ22} & MCM &ChEMBL29 (2M)  & --- & \href{https://github.com/lihan97/kpgt}{Link}\\
&ChemRL-GEM~\cite{Fang2021geo} & Graph + Geometry & GeoGNN~\cite{Fang2021geo} &MCM+CP  & ZINC15 (20M)  & ---&\href{https://github.com/PaddlePaddle/PaddleHelix/tree/dev/apps/pretrained_compound/ChemRL/GEM}{Link} \\
&ImageMol~\cite{zeng2022accurate} & Molecular Images &ResNet18~\cite{DBLP:conf/cvpr/HeZRS16}  & AE + SSC + CP & PubChem ($\sim$10M)  & ---&\href{https://github.com/ChengF-Lab/ImageMol}{Link} \\
&Uni-Mol~\cite{zhou2022uni} & Geometry + Protein Pockets&Transformer&MCM + DN & ZINC/ChemBL + PDB~\cite{berman2000protein} &--- & \href{https://github.com/dptech-corp/Uni-Mol}{Link}\\
\bottomrule
\end{tabular}
\end{adjustbox}
\end{table*}
\paragraph{Multimodal pre-training.}
In addition to the descriptors mentioned in Sec.~\ref{rep}, molecules can also be described using other modalities including images and biochemical texts. Some recent works perform multimodal pre-training on molecules.
For example, KV-PLM~\cite{zeng2022deep} first tokenizes both the SMILES strings and biochemical texts. Then, they randomly mask part of the tokens and pre-train the neural encoders to recover the masked tokens. Analogously, following the replace corrupted spans task of T5~\cite{raffel2020exploring}, MolT5~\cite{edwards2022translation} first masks some spans of abundant SMILES strings and biochemical text descriptions of molecules and then pre-train the transformer to predict the masked spans. In this way, these pre-trained models can generate both the SMILES strings and biochemical texts, which is especially effective for text-guided molecule generation and molecule captioning (generation of the descriptive texts for molecules). \cite{zhu2022featurizations} propose to maximize the consistency between the embeddings of four molecular descriptors and their aggregated embedding using a contrastive objective. In this way, these various descriptors can collaborate with each other for molecular property prediction tasks. Additionally, MICER~\cite{yi2022micer} adopts an autoencoder-based pre-training framework for molecular image captioning. Specifically, they feed molecular images to the pre-trained encoder and then decode the corresponding SMILES strings. The above-mentioned multimodal pre-training strategies can advance the translations between various modalities. Also, these modalities can work together to create a more complete knowledge base for various downstream tasks.

\section{Applications}
\label{app}
The following section takes drug discovery as a case study and showcases several promising applications of CPMs (Tab.~\ref{Table_pGMs}).
\subsection{Molecular Property Prediction (MPP)}
The bioactivity of a new drug candidate is influenced by various factors in real life, including solubility in the gastrointestinal tract, intestinal membrane permeability, and intestinal/hepatic first-pass metabolism. However, such labels for molecules can be extremely scarce because wet-lab experiments are often laborious and expensive. CPMs offer a solution that can exploit the massive unlabeled molecules and serve as powerful backbones for downstream molecular property prediction tasks~\cite{Wang2021MolCLRMC,zaidi2022pre}. Furthermore, compared with the models trained from scratch, CPMs can better extrapolate to out-of-distribution molecules, which is especially important when predicting the properties of newly synthesized drugs~\cite{Hu*2020Strategies}.

\subsection{Molecular Generation (MG)} 
Molecular generation, a long-standing challenge in computer-aided drug design, has been revolutionized by machine learning methods, especially generative models, that narrow the search space and improve computational efficiency, making it possible to delve into the seemingly infinite drug-like chemical space~\cite{du2022molgensurvey}. CPMs, such as MolGPT~\cite{bagal2021molgpt}, which employs an autoregressive pre-training approach, has proven to be instrumental in generating valid, unique, and novel molecular structures. 
The emergence of multi-modal molecular pre-training techniques~\cite{edwards2022translation,zeng2022deep} has further expanded the possibilities of molecular generation by enabling the transformation of descriptive text into molecular structures.
Another crucial area where CPMs have demonstrated their prowess is the generation of three-dimensional molecular conformations, particularly for the prediction of protein-ligand binding poses. Unlike conventional approaches based on molecular dynamics or Markov chain Monte Carlo, which are often hindered by computational limitations, especially for larger molecules~\cite{hawkins2017conformation}, CPMs based on 3D geometry~\cite{zhu2022unified,zhou2022uni} exhibit remarkable superiority in conformation generation tasks, as they can capture some inherent relationships between 2D molecules and 3D conformations during the pre-training process.

\subsection{Drug-Target Interactions (DTI)}
Predictive analysis of Drug-Target Interactions (DTI) is a vital step in the early stages of drug discovery, as it helps to identify drug candidates with binding potential to specific protein targets.
This is particularly important in drug repurposing, where the goal is to recycle approved drugs for a new disease, thereby reducing the need for further drug discovery and minimizing safety risks.
Similarly, obtaining sufficient drug-target data for supervised training can be challenging.
CPMs can overcome this issue by providing molecular encoders with good initializations.
To achieve accurate DTI prediction using CPMs, it is essential to consider both molecular encoders and target encoders, predict binding affinities, and co-train both for the DTI prediction task~\cite{nguyen2021graphdta}.
Previous works such as MPG~\cite{li2021effective} follow these principles to advance DTI predictions.

\subsection{Drug-Drug Interactions (DDI)}
Accurately predicting Drug-Drug Interactions (DDI) is another crucial stage in drug discovery pipelines as such interactions can result in adverse reactions that can harm health and even cause death.
Moreover, accurate DDI predictions can also assist in making informed medication recommendations, making it an essential part of the regulatory investigation prior to market approval.
From the machine learning perspective, DDI prediction can be regarded as a classification task that determines the influence of combination drugs as synergistic, additive, or antagonistic.
To achieve effective DDI prediction, expressive molecular representations are required, which can be obtained using CPMs.
MPG~\cite{li2021effective} is a representative example that has demonstrated the usefulness of CPMs by adopting DDI prediction as a downstream task.
 

\section{Conclusions and Future Outlooks}
In conclusion, this paper provides a comprehensive overview of Chemical Pre-trained Models. We start by reviewing the widely-used descriptors and encoders for molecules, then present the representative pre-training strategies and evaluate their advantages and disadvantages. We also showcase various successful applications of CPMs in drug discovery and development.
Despite the fruitful progress, there are still several challenges that warrant further research in the future.

\subsection{Improving Encoder Architectures and Pre-training Objectives}
While remarkable advancements have been achieved in analyzing the learning capabilities of neural architectures, such as the WL-test for GNNs, these analyses lack specificity in determining the optimal design for highly structured molecules.
The ideal featurizations and architectures for CPMs remain elusive, as evidenced by conflicting results such as the negative impact of Graph Attention Networks (GATs)~\cite{velickovic2018graph}, which are widely adopted in graph learning, on downstream performance in previous studies~\cite{Hu*2020Strategies,hou2022graphmae}.
Furthermore, there is a pressing need to explore ways of seamlessly integrating message-passing techniques into transformers as a unified encoder to accommodate pre-training of large-scale molecular graphs. 
Additionally, as discussed in Sec.~\ref{PS}, the pre-training objectives still leave much room for improvement, with the efficient masking strategy of subcomponents in MCM being a prime example.

\subsection{Building Reliable and Realistic Benchmarks}
Despite the numerous studies conducted on CPMs, their experimental results can sometimes be unreliable due to the inconsistent evaluation settings employed (e.g., random seeds and dataset splits).
For instance, on MoleculeNet~\cite{molnet} that contains several expensive datasets for molecular property prediction, the performance of the same model can vary significantly with different random seeds, possibly due to the relatively small scale of these molecular datasets.
It is also crucial to establish more reliable and realistic benchmarks for CPMs that take out-of-distribution generalization into account.
One solution is to evaluate CPMs through scaffold splitting, which involves splitting molecules based on their substructures.
In reality, researchers must often apply CPMs trained from already known molecules to newly synthesized, unknown molecules that may differ greatly in properties and belong to divergent domains.
In this regard, the recently established Therapeutics Data Commons (TDC)~\cite{huang2021therapeutics} offers a promising opportunity to fairly evaluate CPMs across a diverse range of therapeutic applications.

\subsection{Broadening the Impact of Chemical Pre-trained Models}
The ultimate goal of CPMs studies is to develop versatile molecular encoders that can be applied to a plethora of downstream tasks related to molecules. Nonetheless, compared to the progress of PLMs in the NLP community, there remains a substantial disparity between methodological advancements and practical applications of CPMs. On one hand, the representations produced by CPMs have not yet been extensively used to replace conventional molecular descriptors in chemistry, and the pre-trained models have not yet become standard tools for the community. On the other hand, there is limited exploration into how these models can benefit a more extensive range of downstream tasks beyond individual molecules, such as chemical reaction prediction, molecular similarity searches in virtual screening, retrosynthesis, chemical space exploration, and many others.

\subsection{Establishing Theoretical Foundations}
Even though CPMs have demonstrated impressive performance in various downstream tasks, a rigorous theoretical understanding of these models has been limited. This lack of underpinnings presents a hindrance to both the scientific community and industry stakeholders who seek to maximize the potential of these models.
The theoretical foundations of CPMs must be established in order to fully comprehend their mechanisms and how they drive improved performance in various applications.
For example, a recent empirical study~\cite{sun2022does} has questioned the superiority of certain self-supervised graph pre-training strategies over non-pre-trained counterparts in some downstream tasks.
Further research is necessary to gain a more robust understanding of the effectiveness of different molecular pre-training objectives, so as to provide guidance for optimal methodology design.

\begin{spacing}{1}
{
\small
\bibliographystyle{named}
\bibliography{ijcai23}
}
\end{spacing}

\end{document}